\documentclass[twoside]{article}
\usepackage{xspace}
\usepackage{ecj,palatino,epsfig,latexsym,natbib}
\usepackage{array, algpseudocode}
\usepackage{amsmath, amssymb, amsfonts, parskip, graphicx, verbatim}
\usepackage{hyperref}
\usepackage{pgf, tikz, microtype, algorithm}
\newcommand{\IOH}{{IOHexperimenter}\xspace}

\begin{document}

\ecjHeader{x}{x}{xxx-xxx}{2021}{Benchmarking with IOHexperimenter}{J. de Nobel, F. Ye, D. Vermetten, H. Wang, C. Doerr, T. B\"ack}
\title{\bf IOHexperimenter: Benchmarking Platform for Iterative Optimization Heuristics}  

\author{\name{\bf Jacob de Nobel}\footnotemark[1] \hfill \addr{
j.p.de.nobel@liacs.leidenuniv.nl}
\AND
       \name{\bf Furong Ye\thanks{\hspace{7pt}These authors contributed equally to this work.}} \hfill \addr{f.ye@liacs.leidenuniv.nl}
\AND
       \name{\bf Diederick Vermetten} \hfill \addr{
d.l.vermetten@liacs.leidenuniv.nl}
\AND
       \name{\bf Hao Wang} \hfill \addr{h.wang@liacs.leidenuniv.nl}
    \\ \addr{LIACS, Leiden University, the Netherlands}
\AND
       \name{\bf Carola Doerr} \hfill \addr{Carola.Doerr@lip6.fr}\\
        \addr{LIP6, Sorbonne Universit\'e, Paris, France}
\AND
       \name{\bf Thomas B\"ack} \hfill \addr{
t.h.w.baeck@liacs.leidenuniv.nl}
       \\ \addr{LIACS, Leiden University, the Netherlands}
}

\maketitle

\begin{abstract}
\vspace{10pt}
We present IOHexperimenter, the experimentation module of the IOHprofiler project, which aims at providing an easy-to-use and highly customizable toolbox for benchmarking iterative optimization heuristics such as local search, evolutionary and genetic algorithms, Bayesian optimization techniques, etc. IOHexperimenter can be used as a stand-alone tool or as part of a benchmarking pipeline that uses other components of IOHprofiler such as IOHanalyzer, the module for interactive performance analysis and visualization.

IOHexperimenter provides an efficient interface between optimization problems and their solvers while allowing for granular logging of the optimization process. These logs are fully compatible with existing tools for interactive data analysis, which significantly speeds up the deployment of a benchmarking pipeline. The main components of IOHexperimenter are the environment to build customized problem suites and the various  logging options that allow users to steer the granularity of the data records. 
\end{abstract}

\begin{keywords}
\vspace{10pt}

Iterative Optimization Heuristics, 
Benchmarking,
Algorithm Comparison

\end{keywords}

\section{Introduction}

In order to compare and to improve upon state-of-the-art optimization algorithms, it is important to be able to gain insights into their search behavior on wide ranges of problems. To do so systematically, a robust benchmarking setup has to be created that allows for rigorous testing of algorithms. 
While many benchmarking projects exist~\citep{abs-2007-03488,paradiseo_logger}, they are often created for specific research problems and hard to extend beyond their original scope. To address this limitation, we have designed the benchmarking environment, IOHexperimenter, which strongly emphasizes extensibility and customizability, allowing users to add new problems or to build interfaces with other benchmarking software. IOHexperimenter supports customized logging of algorithm performance. The extendability of IOHexperimenter distinguishes it from other well-known benchmark projects such as COCO~\citep{hansen2021coco}, which cannot be easily extended to include new benchmark problems, and Nevergrad~\citep{nevergrad}, which provides a platform for comparing built-in optimizers but not customized algorithms and which does not provide logging functionality beyond fixed-budget simple regret.

\qquad 

\IOH is a part of the overarching IOHprofiler project, which connects algorithm frameworks, problem suites, interactive data analysis, and performance repositories together in an extendable benchmarking pipeline. Within this pipeline, \IOH can be considered the interface between algorithms and problems, where it allows consistent collection of performance data and of algorithmic data such as the evolution of control parameters that change during the optimization process. 

\qquad 
To perform the benchmarking, three component interact with each other: \emph{problems}, \emph{loggers}, and \emph{algorithms}.
Within \IOH, an interface is provided to ensure that any of these components can be modified without impacting the behavior of the others, in the sense that any changes to their setup will be compatible with the other components of the benchmarking pipeline.

\begin{figure*}
    \vspace{-2mm}
    \centering
    \includegraphics[width=0.8\linewidth]{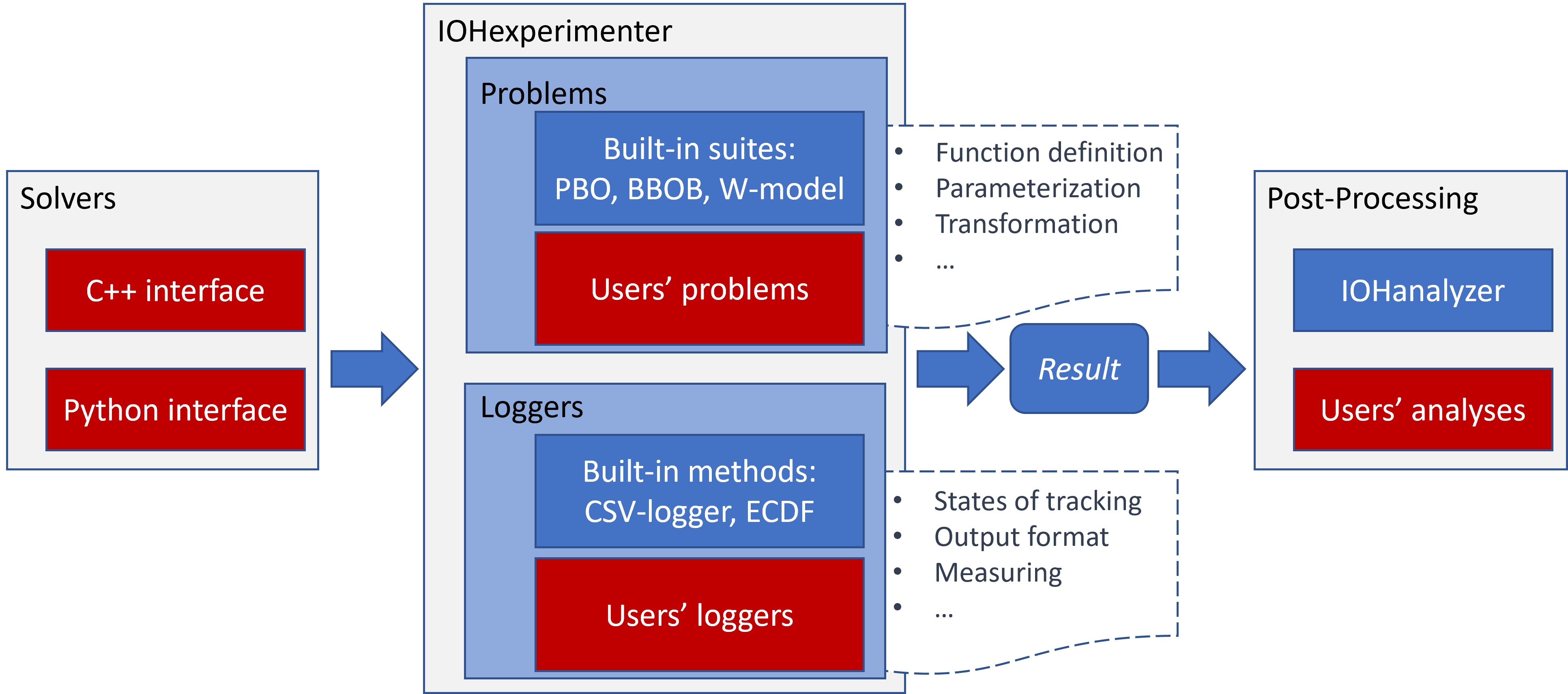}
    \caption{Workflow of \IOH
    }
    \label{fig:IOH}
    \vspace{-6mm}
\end{figure*}

\section{Functionality}
At its core, \IOH provides a standard interface towards expandable benchmark \emph{problems} and several \emph{loggers} to track the performance and the behavior (internal parameters and states) of \emph{algorithms} during the optimization process.
The logger is integrated into a wide range of existing tools for benchmarking, including \emph{problem} suites such as PBO~\citep{doerr2020benchmarking} and the W-model~\citep{weise2018difficult} 
for discrete optimization and COCO's noiseless real-valued single-objective BBOB problems~\citep{hansen2021coco} for the continuous case.
On the \emph{algorithms} side, \IOH has been connected to several algorithm frameworks, including ParadisEO~\citep{paradiseo_logger}, a modular genetic algorithm~\citep{modga}, a modular CMA-ES~\citep{modcma}, and the optimizers in Nevergrad~\citep{nevergrad}.
The output generated by the included \emph{loggers} is compatible with the IOHanalyzer module~\citep{iohana} for interactive performance analysis.

\qquad
Figure~\ref{fig:IOH} shows the way \IOH can be placed in a typical benchmarking workflow. The key factor here is the flexibility of its design: \IOH can be used with any user-provided solvers and problems given a minimal overhead, and ensures output of experimental results which follow conventional standards. Because of this, the data produced by \IOH is compatible with post-processing frameworks like IOHanalyzer~\cite{iohana}, enabling an efficient path from algorithm design to performance analysis. In addition to the built-in interfaces to existing software, \IOH aims at providing a user-friendly, easily accessible way to customize the benchmarking setup.
\IOH is built in C++, with an interface to Python. In this paper, we describe the functionality of the package on a high level, without going into implementation details.\footnote{Technical documentation for both C++ and Python can be found on the IOHprofiler wiki at \url{https://iohprofiler.github.io/}, which provides a getting-started and several use-cases.} 
In the following, we introduce the typical usage of \IOH, as well as the ways in which it can be customized to fit different benchmarking scenarios.

\subsection{Problems}
In \IOH, a problem instance is defined as $F = T_y \circ f \circ T_x$, in which $f\colon X \rightarrow \mathbb{R}$ is a benchmark problem (e.g., for \textsc{OneMax} $X=\{0,1\}^n$ and for the sphere function $X=\mathbb{R}^n$), and $T_x$ and $T_y$ are automorphisms supported on $X$ and $\mathbb{R}$, respectively, representing transformations in the problem's domain and range (e.g., translations and rotations for $X=\mathbb{R}^n$). To generate a problem instance, one needs to specify a tuple of a problem $f$, an instance identifier $i\in\mathbb{N}_{>0}$, and the dimension $n$ of the problem. 
Any problem instances that reconcile with this definition of $F$, can easily be integrated into \IOH, using the C++ core or the Python interface.\footnote{Note that multi-objective problems do not follow this structure, and are not yet supported within \IOH. Integration of both noisy and mixed-variable type objective functions is in development.}

\qquad
The transformation methods are particularly important for robust benchmarking, as they allow for the creation of multiple problem instances from the same base-function. They allow the user to check algorithm invariance to search space transformations, such as scaling, rotation, and translation. Built-in transformations for pseudo-Boolean functions are available~\citep{IOHprofiler}, as well as transformation methods for continuous optimization used by~\citep{hansen2021coco}. Additionally, problems can be combined in a \emph{suite}, which allows the solver to easily run on the selected problem instances.

\subsection{Data Logging}
\IOH provides \emph{loggers} to track the performance of algorithms during the optimization process. These \emph{loggers} can be tightly coupled with the problems: when evaluating a solution, the attached loggers will be triggered to store relevant information. information about solution quality is always recorded, while algorithm's control parameters are included only if specified by the user. The events that trigger a data record are customized by the user; e.g., via specifying a frequency at which information is stored, or by choosing quality thresholds that trigger a data record when they are met for the first time.

\qquad
The default logger makes use of a two-part data format: meta-information such as function id, instance, and dimension, which gets written to \verb|.info|-files, and the performance data that gets written to space-separated \verb|.dat|-files. A full specification of this format can be found in~\cite{iohana}. Data in this format can be used directly with the IOHanalyzer module.

In addition to the built-in loggers, custom logging functionality can be created within \IOH as well.
For example, a logger storing only the final calculated performance measure was created for algorithm configuration tasks~\citep{paradiseo_logger}.

\section{Conclusions and Future Work}

IOHexperimenter is a tool for benchmarking iterative optimization heuristics. It aims at making rigorous benchmarking more approachable by providing a structured benchmarking pipeline which can be adapted to fit a wide range of scenarios. The combination of a clear output format and common interface across both Python and C++ makes IOHexperimenter a useful component towards reproducible algorithm comparison.

While currently \IOH only supports single-objective and noiseless optimization, an extension to other types of problems is desirable, allowing for more general usage of the IOHexperimenter. Additionally, support for arbitrary combinations of variable types would enable the creation of benchmark suites in the mixed-integer optimization domain.

IOHexperimenter can be slotted into a benchmarking pipeline by generating output data for the IOHanalyzer module, which provides a highly interactive analysis of algorithms performance.
Its customized logging functionality allows \IOH to be used in machine learning scenarios such as algorithm configuration. 

The IOHprofiler project welcomes contributions of problems from various domains and loggers with different perspectives. We appreciate feedback and comments via \href{mailto: iohprofiler@liacs.leidenuniv.nl}{iohprofiler@liacs.leidenuniv.nl}.

\bibliographystyle{apalike}
\bibliography{ecjsample}

\end{document}